\documentclass[review]{elsarticle}

\usepackage{lineno,hyperref}
\modulolinenumbers[5]
\usepackage{geometry}
\usepackage{natbib}
\usepackage{amssymb}
\usepackage{amsmath}
\usepackage{algorithmic}
\usepackage{algorithm}
\usepackage{enumitem}
\usepackage{enumerate}
\usepackage{geometry}
\usepackage{amsmath,bm}
\biboptions{numbers,sort&compress}
\usepackage{amssymb}
\usepackage{indentfirst}
\usepackage{color}
\usepackage{subfigure}
\usepackage{multirow}
\journal{International Journal of Interactive Multimedia and Artificial Intelligence}

\begin{document}

\begin{frontmatter}

\author{Yongfa Ling $^{\rm 1}$, Wenbo Guan $^{\rm 2}$, Qiang Ruan $^{\rm 3}$, Heping Song $^{\rm 4*}$, Yuping Lai $^{\rm 5}$}
\address{$^{\rm 1}$School of Artificial Intelligence, Hezhou University, Hezhou, China}
\address{$^{\rm 2}$School of Information Science and Technology, North China University of Technology, Beijing, China}
\address{$^{\rm 3}$DigApis Information Security Technology Co., Ltd., Nantong, Jiangsu, China}
\address{$^{\rm 4}$School of Computer Science and Telecommunications Engineering, Jiangsu University, Zhenjiang, China}
\address{$^{\rm 5}$School of Cyberspace Security, Beijing University of Posts and Telecommunications, Beijing, China}
\cortext[mycorrespondingauthor]{Corresponding author (email: songhp@ujs.edu.cn)}

\title{Variational Learning for the Inverted Beta-Liouville Mixture Model and Its Application to Text Categorization}

\begin{abstract}
The finite invert Beta-Liouville mixture model (IBLMM) has recently gained some attention due to its  positive data modeling capability.
Under the conventional variational inference (VI) framework, the analytically tractable solution to the optimization of the variational posterior distribution cannot be obtained, since the variational object function involves evaluation of intractable moments. With the recently proposed extended variational inference (EVI) framework, a new function is proposed to replace the original variational object function in order to avoid intractable moment computation, so that the analytically tractable solution of the IBLMM can be derived in an elegant way. The good performance of the proposed approach is demonstrated by experiments with both synthesized data and a real-world application namely text categorization.
\end{abstract}


\end{frontmatter}

\linenumbers

\section{Introduction}
Positive data arise naturally in many real-world applications, such as object clustering \cite{Mashrgy2014Robust}, scene categorization \cite{Bdiri2016Variational}, image segmentation \cite{Liu2019Bayesian}, and object detection \cite{Lai2021Extended}. During the last decade, many non-Gaussian mixture models,  \emph{e.g.}, the finite inverted Dirichlet mixture model (IDMM) \cite{Bdiri2012Positive,Yuping2018Variational}, the finite generalized inverted Dirichlet mixture model (GIDMM) \cite{Bourouis2014Bayesian}, the finite generalized Gamma mixture model (GGaMM) \cite{Liu2019Bayesian} and the finite inverted Beta-Liouville mixture model (IBLMM) \cite{Hu2019A}, were proposed to model and analyze positive data due to their powerful modeling capabilities. Among these mixture models, the IBLMM is one of the most popular approaches for modeling univariate and multivariate positive data. For example, the IBLMM is shown to be very flexible and powerful in analyzing and clustering text documents \cite{Hu2019A}, therefore, modeling positive data with the IBLMM is well-motivated.

The major task in modeling the data with the finite mixture models is the learning of the model parameters, which refers to both estimating the model parameters and determining the number of components (\emph{i.e.}, the model complexity). A variety of approaches can be applied to address this problem, such as the expectation maximization (EM) algorithm \cite{Agusta2003Unsupervised}, the Markov chain Monte Carlo (MCMC) \cite{Favaro12013MCMC}, the expectation propagation (EP) \cite{Thomas2001Expectation} and the variational inference (VI) \cite{Bishop2006Pattern}. Among these approaches, the VI has been the most popular method. Much of its popularity is due to the fact that it may scale well to large applications. The main idea behind the VI is to find a approximate distribution for the intractable real posterior distribution by minimizing the Kullback-Leibler (KL) divergence of these two distributions. This is equivalent to maximizing the evidence lower bound (ELBO), which is  also known as the variational objective function. Unfortunately, it is infeasible to obtain an analytical solution to the VI for many non-Gaussian mixtures, such as the IDMM, the GIDMM, the GGaMM and the IBLMM,  since some computationally intractable moments exist in the ELBO. This problem can be elegantly solved by the recently proposed extended variational inference (EVI) \cite{Ma2019Insights}. The main idea behind the EVI framework is that the optimal solutions can be obtained by means of maximizing a lower bound of the ELBO. This bound can be obtained by introducing some tractable approximations to the original objective function.

Motivated by the powerful modeling capability of the IBLMM and the excellent performance achieved by the EVI framework, the EVI framework is applied to learn the IBLMM. The major contributions of this work can be summarized as follows. First, the analytical solution within the EVI framework for the IBLMM is derived. In this framework, the estimated values of all the involved parameters and the number of components can be simultaneously obtained. Second, the proposed approach is used in an important real-world application namely text categorization. Synthesized and real data evaluations demonstrate the good performance of the model trained by the proposed approach.

The reminder of this paper is organized as follows. In Section 2, a brief review of the IBLMM is given. In Section 3, the Bayesian learning algorithm with the EVI is derived. The experimental results on synthesized and real datasets are reported in Section 4. Finally, some conclusions are drawn in Section 5.

\section{Preliminaries}
A brief overview of the IBLMM is given first in this section. Then, a complete Bayesian framework for this model is presented.
\subsection{Finite Inverted Beta-Liouville mixture model}
If a $D$-dimensional random vector $\mathbf x={{[{{x}_{1}},\cdots,{{x}_{D}}]}^{\text T}}$ contains positive values, the underlying distribution of $\mathbf x$ can be modeled by the inverted Beta-Liouville (IBL) distribution. The probability density function (PDF) of the IBL distribution is given by \cite{Fan2020Modeling}
\begin{equation}\label{eq:1}
\begin{split}
p({\mathbf{x}}|\bm\alpha,u,v){\rm{ = }}&\frac{{\Gamma (\sum\nolimits_{d = 1}^D {{\alpha _d}} )\Gamma (u + v)}}{{\Gamma (u )\Gamma ( v)}}\prod\limits_{d = 1}^D {\frac{{x_d^{{{\alpha _d}-1}}}}{{\Gamma ({\alpha _d})}}}{\left( {\sum\limits_{d = 1}^D {{x_d}} } \right)^{u - \sum\nolimits_{d = 1}^D {{\alpha _d}} }}{\left( {1 + \sum\limits_{d = 1}^D {{x_d}} } \right)^{ - (u + v)}},
\end{split}
\end{equation}
where $\bm\alpha ={{[{{\alpha }_{1}},\cdots ,{{\alpha }_{D}}]}^{\text T}}$, $\Gamma (\cdot )$ is the Gamma function defined as
$\Gamma (a) = \int_0^\infty  {{t^{a - 1}}} {e^{ - t}}dt$.

To model the multimodality  of the observed data $\mathbf X=[{{\mathbf x}_{1}},\cdots ,{{\mathbf x}_{N}}]$, the mixture modeling
technique \cite{McLachlan2000Finite} is used to construct the IBLMM with the PDF as follows
\begin{equation}\label{eq:2}
p(\mathbf X|\bm\Lambda ,{\bf{u}},{\bf{v}},\bm\pi ) = \prod\limits_{n = 1}^N  {\sum\limits_{m = 1}^M {\pi_m} {p({{\bf{x}}_{\bf{n}}}|{\bm\alpha _m},{u_m},{v_m})} },
\end{equation}
where $M$ is the number of components, $\bm\pi ={{[{{\pi }_{m}},\cdots ,{{\pi }_{M}}]}^{\text T}}$ is the mixing weights, $\bm\Lambda \text{=}[{{\bm\alpha }_{1}},\cdots ,{{\bm\alpha }_{M}}]$, ${\bf{u}}{\rm{ = }}{[{u_1}, \cdots ,{u_M}]^{\text T}}$ and ${\bf{v}}{\rm{ = }}{[{v_1}, \cdots ,{v_M}]^{\text T}}$ denote the parameter matrices.\\

\subsection{Bayesian Framework for IBLMM}
It is convenient to turn the mixture model in (2) into a latent variable model. For each vector $\mathbf x_{n}$, a latent vector variable  ${{\mathbf {z}}_{n}}\text{=}{{[{{z}_{n1}},\cdots ,{{z}_{nM}}]}^{\text T}}$ is assigned, such that ${{\rm{z}}_{nm}} \in \{ 0,1\}$, $\sum\nolimits_{m=1}^{M}{{{z}_{nm}}=1}$ and ${{z}_{nm}}=1$ if $\mathbf x_{n}$ is drawn from the ${m}$th component and 0 otherwise. 
Then, the latent variable model of IBLMM can be written as
\begin{equation}\label{eq:3}
p(\mathbf Z|\bm\pi )=\prod\limits_{n=1}^{N}{\prod\limits_{m=1}^{M}{\pi _{m}^{{{\text z}_{nm}}}}},
\end{equation}
\begin{equation}\label{eq:4}
p({\mathbf{X}},{\mathbf{Z}}|{\bm{\Lambda}},{\mathbf{u}},{\mathbf{v}} ) = \prod\limits_{n = 1}^N {\prod\limits_{m = 1}^M {p{{({{\bf{x}}_n}|{\bm\alpha _m},{u_m},{v_m})}^{{z_{nm}}}}} },
\end{equation}
where ${\mathbf Z} = {[{\mathbf z_1}, \cdots ,{\mathbf z_M}]^{\text T}}$.

To formulate a full Bayesian mixture model, the conjugate priors on parameters $\bm\Lambda$, $\mathbf u$, $\mathbf v$, and $\bm\pi$ have to be designated as follows:
\begin{equation}\label{eq:5}
p(\bm\Lambda ) = {\cal G}(\bm\Lambda |{\bm g},{\bm h}) = \prod\limits_{m = 1}^M {\prod\limits_{d = 1}^D {\mathcal G({\alpha _{md}}|{g_{md}},{h_{md}})} },
\end{equation}
\begin{equation}\label{eq:6}
p(\mathbf u )=\bm{\mathcal G}(\mathbf u |\bm{s},\bm{t})=\prod\limits_{m = 1}^M {{\cal G}({u_m}|{s_{{m_{\rm{}}}}},{t_{{m_{\rm{}}}}})},
\end{equation}
\begin{equation}\label{eq:7}
p(\mathbf v )=\bm{\mathcal G}(\mathbf v |\bm{p},\bm{q})=\prod\limits_{m = 1}^M {{\cal G}({v_m}|{p_{{m_{\rm{}}}}},{q_{{m_{\rm{}}}}})},
\end{equation}
\begin{equation}\label{eq:8}
p(\bm\pi)=Dir(\bm\pi |\bm c)=\frac{\Gamma (\sum\nolimits_{m=1}^{M}{{{c}_{m}}})}{\prod\nolimits_{m=1}^{M}{\Gamma ({{c}_{m}})}}\prod\limits_{m=1}^{M}{\pi _{m}^{{{c}_{m}}-1}},
\end{equation}

where $\bm g=\{{{g}_{md}}\}$, $\bm h=\{{{h}_{md}}\}$, $\bm s=\{{{s}_{m}}\}$, $\bm t=\{{{t}_{m}}\}$, $\bm p=\{{{p}_{m}}\}$, $\bm q=\{{{q}_{m}}\}$, $\bm c=\{{{c}_{m}}\}$, $\mathcal G( \cdot )$ and $Dir( \cdot )$  denote the Gamma distribution and the Dirichlet distribution, respectively.

Following the Bayes' theorem and combining (3), (4), (5), (6), (7) and (8),  the joint distribution of the observation $\mathbf X$ and all the random variables $\bm\Theta = \{ {\mathbf{Z}},\bm\Lambda ,{\mathbf{u}},{\mathbf{v}},\bm\pi \}$  is given by:
\begin{equation}\label{eq:9}
p(\mathbf{X},\bm\Theta) = p({\mathbf{X}},{\mathbf{Z}}|{\bm{\Lambda}},{\mathbf{u}},{\mathbf{v}})p(\mathbf Z|\bm\pi )p(\bm\pi)p(\bm\Theta)p(\mathbf u )p(\mathbf v ).
\end{equation}

\section{Learning the Model}
\subsection{Extended Variational Inference}
The VI framework \cite{Bishop2006Pattern} is commonly employed to estimate the parameters and determine the optimal number of components of the mixture models.  The major goal is to find an approximate distribution $q(\bm\Theta)$ for the true posterior distribution $p(\bm\Theta|\mathbf X)$. The optimal $q(\bm\Theta)$ can be obtained by maximizing the ELBO as follows:
\begin{equation}\label{eq:10}
\mathcal{L}(q)={\langle \ln p({\bf{X}},\bm\Theta)\rangle _q} - {\langle \ln q(\bm\Theta )\rangle _q},
\end{equation}
where ${\langle  \cdot \rangle _q}$ denotes the expectation regarding the distribution $q$. Note that the $\mathcal L(q)$ is not analytically tractable for most of the non-Gaussian mixture models, such as the IDMM, the GIDMM, the GGaMM and the IBLMM, as (9) involves intractable moments. The recently proposed EVI framework \cite{Ma2019Insights} offers an elegant way to address this problem. The main idea behind
the EVI framework  is that if a ``helping function" $\tilde{p}(\mathbf{X},\bm\Theta)$, which satisfies the constraint ${{\text E}_{q}}[\ln p(\mathbf{X},\bm\Theta)] \ge {{\text E}_{q}}[\ln \tilde p(\mathbf{X},\bm\Theta)]$, can be found, then  the optimal solutions can be reached asymptotically through maximizing a lower bound of the $\mathcal {L}(q)$. This bound is given by
\begin{equation}\label{eq:11}
\mathcal{ L}(q) \ge \mathcal{\tilde L}(q)={{\text E}_{q}}[\ln \tilde p(\mathbf{X},\bm\Theta)]-{{\text E}_{q}}[q(\bm\Theta )].
\end{equation}
To formulate a computationally tractable expression for the $\mathcal {\tilde L}(q)$, the simplest approach called the mean-field approach is adopted which factorizes the $q(\bm\Theta )$ as follows
\begin{equation}\label{eq:12}
\begin{array}{l}
q(\bm\Theta ) = \prod\limits_{n = 1}^N {\prod\limits_{m = 1}^M {q({z_{nm}})} } \prod\limits_{m = 1}^M {\prod\limits_{d = 1}^D {q({\alpha _{md}})} } \prod\limits_{m = 1}^M {\left[ {q({u_m})q({v_m})}q(\pi_{m}) \right]}.\\
\end{array}
\end{equation}
Then, the optimal form of $q(\Theta_{k})$, denoted by $q^{*}(\Theta_{k})$ in this case, is given by
\begin{equation}\label{eq:13}
\ln q_{k}^{*}({{\Theta }_{k}})={{\langle \ln \tilde{p}(\mathbf{X},\bm\Theta \text{ }\!\!\!\!\text{ })\rangle }_{s\ne k}}+ \text{Cst},
\end{equation}
where ${{\langle \cdot \rangle }_{s\ne k}}$ denotes the expectation regards all factors $q_{s}(\Theta_{s})$ except for $s=k$ and ``Cst'' denotes a normalizing constant. In the EVI framework, all factors $q_{s}(\Theta_{s})$ are need to be initiate first and then each factor is updated by updating the hyper-parameters.
\subsection{Variational Distribution}
This section details how \eqref{eq:13} is applied to compute the variational factors. Note that the EVI is essentially iterative, since it represents a distribution factor applying knowledge about other factors. Following the principles of the EVI framework, the expectation of the joint distribution's logarithm is first calculated as
\begin{equation}\label{eq:14}
\begin{split}
\langle \ln p({\bf{X}},\bm\Theta)\rangle  = &\sum\limits_{n = 1}^N {\sum\limits_{m = 1}^M {\langle {z_{nm}}\rangle \left\{ {\langle\ln {\pi _m}\rangle + {\mathcal R_{m}} + {\mathcal F_m}} \right.} } \; + \sum\limits_{d = 1}^D {(\langle } {\alpha _{md}}\rangle {\rm{ - 1}})\ln {x_{nd}}\\
&+ \ln (\sum\limits_{d = 1}^D {{x_{nd}}} )(\langle {u_m}\rangle  - \sum\limits_{d = 1}^D {\langle {\alpha _{md}}\rangle } )\left. { - (\langle {u_m}\rangle  + \langle {v_m}\rangle )\ln (1 + \sum\limits_{d = 1}^D {{x_{nd}}} )} \right\}\;\\
&+ \sum\limits_{m = 1}^M {\sum\limits_{d = 1}^D {\left[ {({g_{md}} - 1)} \right.} } \langle \ln {\alpha _{md}}\rangle \left. { - {h_{md}}\langle {\alpha _{md}}\rangle } \right]\; + \sum\limits_{m = 1}^M {[({s_m} - 1)} \langle \ln {u_m}\rangle \left. { - {t_m}\langle {u_m}\rangle } \right]\;\\
& + \sum\limits_{m = 1}^M {[({p_m} - 1)} \langle \ln {v_m}\rangle \left. { - {q_m}\langle {v_m}\rangle } \right]\; + \sum\limits_{m = 1}^M {({c_m} - 1)\langle\ln {\pi _m}} \rangle+ {\rm{Cst}},
\end{split}
\end{equation}
where ${{\cal R}_{m}} = \left\langle {\ln \frac{{\Gamma (\sum\nolimits_{d = 1}^D {{\alpha _{md}}} )}}{{\prod\nolimits_{d = 1}^D {\Gamma ({\alpha _{md}})} }}} \right\rangle$, ${{\cal F}_m} = \left\langle {\ln \frac{{\Gamma ({u_m} + {v_m})}}{{\Gamma ({u_m})\Gamma ({v_m})}}} \right\rangle$. It is noteworthy that \eqref{eq:14} is not available in a closed form because it includes the intractable moments ${{\cal R}_{m}}$, ${{\cal F}_{m}}$. Following the principles of the aforementioned EVI framework, two ``helping functions" $\tilde{\mathcal R}_{m}$, $\tilde{\mathcal F}_{m}$, satisfying $\mathcal R_{m}\ge \tilde{\mathcal R}_{m}$, $\mathcal F_{m}\ge \tilde{\mathcal F}_{m}$, respectively have to be found. According to \cite{Ma2014Bayesian}, $\tilde{\mathcal R}_{m}$ and $\tilde{\mathcal F}_{m}$ is obtained as follows:
\begin{equation}\label{eq:15}
\begin{array}{l}
{{\tilde {\cal R}}_m} = \ln \frac{{\Gamma (\sum\nolimits_{d = 1}^D {{{\bar \alpha }_{md}}} )}}{{\prod\nolimits_{d = 1}^D {\Gamma ({{\bar \alpha }_{md}})} }} + \sum\limits_{d = 1}^D {\left[ {\Psi (\sum\limits_{k = 1}^D {{{\bar \alpha }_{mk}}} ) - \Psi ({{\bar \alpha }_{md}})} \right]} \left[ {\langle \ln {\alpha _{md}}\rangle - \ln {{\bar \alpha }_{md}}} \right]{{\bar \alpha }_{md}},
\end{array}
\end{equation}
\begin{equation}\label{eq:16}
\begin{split}
{{\tilde{\cal F}}_m} = & \ln \frac{{\Gamma ({{\bar u}_m} + {{\bar v}_m})}}{{\Gamma ({{\bar u}_m})\Gamma ({{\bar v}_m})}} + [\Psi ({{\bar u}_m} + {{\bar v}_m}) - \Psi ({{\bar u}_m})](\langle \ln {u_m}\rangle - \ln {{\bar u}_m}){{\bar u}_m} \\
& + [\Psi ({{\bar u}_m} + {{\bar v}_m}) - \Psi ({{\bar v}_m})](\langle \ln {v_m}\rangle - \ln {{\bar v}_m}){{\bar v}_m},
\end{split}
\end{equation}
where
\begin{equation}\label{eq:17}
 {{\bar \alpha }_{md}} = \langle {\alpha _{md}}\rangle, {{\bar u}_m} = \langle {u_m}\rangle, {{\bar v}_m} = \langle {v_m}\rangle, \Psi (a) = \frac{{\partial \ln \Gamma (a)}}{{\partial a}}.
\end{equation}

Insert (15) and (16) into (14) then a lower bound to $\langle \text{ln}p(\mathbf{X},\bm\Theta)\rangle $ is obtained as
\begin{equation}\label{eq:18}
\begin{split}
\langle \ln \tilde p({\bf{X}},\bm\Theta)\rangle=  & \sum\limits_{n = 1}^N {\sum\limits_{m = 1}^M {\langle {z_{nm}}\rangle \left\{ {\langle\ln {\pi _m}\rangle + {\tilde{\mathcal R}_{m}} + {\tilde{\mathcal F}_m}} \right.} } \; + \sum\limits_{d = 1}^D {(\langle } {\alpha _{md}}\rangle {\rm{ - 1}})\ln {x_{nd}}\\
&+ \ln (\sum\limits_{d = 1}^D {{x_{nd}}} )(\langle {u_m}\rangle  - \sum\limits_{d = 1}^D {\langle {\alpha _{md}}\rangle } )\left. { - (\langle {u_m}\rangle  + \langle {v_m}\rangle )\ln (1 + \sum\limits_{d = 1}^D {{x_{nd}}} )} \right\}\;\\
&+ \sum\limits_{m = 1}^M {\sum\limits_{d = 1}^D {\left[ {({g_{md}} - 1)} \right.} } \langle \ln {\alpha _{md}}\rangle \left. { - {h_{md}}\langle {\alpha _{md}}\rangle } \right]\; + \sum\limits_{m = 1}^M {[({s_m} - 1)} \langle \ln {u_m}\rangle \left. { - {t_m}\langle {u_m}\rangle } \right]\;\\
&+ \sum\limits_{m = 1}^M {[({p_m} - 1)} \langle \ln {v_m}\rangle \left. { - {q_m}\langle {v_m}\rangle } \right]\; + \sum\limits_{m = 1}^M {({c_m} - 1)\langle\ln {\pi _m}}\rangle + {\rm{Cst}}.
\end{split}
\end{equation}
Now, $\bm\alpha$, $\bf u$, and $\bf v$ are the $i.i.d$ variables. Details about solving the optimal variational factors using \eqref{eq:13} is given as follows.

1) $q^{*}(\mathbf Z)$: Including all terms that do not depend upon $z_{nm}$ into a constant term, the equation (19) is obtained as follows
\begin{equation}\label{eq:19}
\ln q^{*}({ z_{nm}}) = \sum\limits_{n = 1}^N {\sum\limits_{m = 1}^M {{z_{nm}}\ln {\rho _{nm}}} } + \text{Cst},
\end{equation}
where
\begin{equation}\label{eq:20}
\begin{array}{l}
\ln {\rho _{nm}} = \ln {\pi _m} + {\tilde{\mathcal R}_m} + {\tilde{\mathcal F}_m} + \sum\limits_{d = 1}^D {({{\bar \alpha }_{md}} - 1)\ln {x_{nd}}} + ({\bar u_m} - \sum\limits_{d = 1}^D {{{\bar \alpha }_{md}}} )\;\ln (\sum\limits_{d = 1}^D {{x_{nd}}} )\\
\;\;\;\;\;\;\;\;\;\;\;\;\; \; \; - ({\bar u_m} +  {\bar v_m} )\ln (1 + \sum\limits_{d = 1}^D {{x_{nd}}} ).
\end{array}
\end{equation}
Taking  exponential of both sides of \eqref{eq:19}, $q^{*}(\mathbf Z)$ is recognized to be a categorical density
\begin{equation}\label{eq:21}
q^{*}(\mathbf Z)=\prod\limits_{n=1}^{N}{\prod\limits_{m=1}^{M}{r_{nm}^{{{z}_{nm}}}}},
\end{equation}
where
\begin{equation}\label{eq:22}
{{r}_{nm}}=\frac{{{\rho }_{nm}}}{\sum\nolimits_{m=1}^{M}{{{\rho }_{nm}}}},
\end{equation}
where $r_{nm}$ are nonnegative and have a unit sum.

\emph{2)} $q^{*}(\bm\Lambda)$: Absorbing any terms independent of $\alpha_{md}$ into the
additive constant results in
\begin{equation}\label{eq:23}
\ln q^{*}({\alpha _{md}}) = (g_{md}^* - 1)\ln {\alpha _{md}} - h_{md}^*{\alpha _{md}} + \text{Cst},
\end{equation}
where $g_{md}^* $ and $h_{md}^* $ are defined by
\begin{equation}\label{eq:24}
g_{md}^* = {g_{md}} + [\Psi (\sum\limits_{k = 1}^D {{\bar\alpha _{md}}} ) - \Psi ({{\bar \alpha }_{md}})]{{\bar \alpha }_{md}}\sum\limits_{n = 1}^N {\langle {z_{nm}}\rangle },
\end{equation}
\begin{equation}\label{eq:25}
h_{md}^* = {h_{md}} - \sum\limits_{n = 1}^N {\langle {z_{nm}}\rangle [\ln {x_{nd}} - \ln (\sum\limits_{d = 1}^D {{x_{nd}}} )]}.
\end{equation}
Taking the exponential of both sides of \eqref{eq:23}, the equation (26) is obtained as follows
\begin{equation}\label{eq:26}
q^{*}(\bm\Lambda ) = \prod\limits_{m = 1}^M {\prod\limits_{d = 1}^D {\mathcal G({\alpha _{md}}|g_{md}^*,h_{md}^*)}}.
\end{equation}

\emph{3)} $q^{*}(\bf u)$: Any terms which are independent of  $u_{m}$ will be absorbed into the additive constant as
\begin{equation}\label{eq:27}
\ln q^{*}({u _{m}}) = (s_{m}^* - 1)\ln {u _{m}} - t_{md}^*{u_{m}} + \text{Cst},
\end{equation}
where $s_m^* $ and $t_m^* $ are given by
\begin{equation}\label{eq:28}
s_m^* = {s_m} + [\Psi ({{\bar u}_m} + {{\bar v}_m}) - \Psi ({{\bar u}_m})]{{\bar u}_m}\sum\limits_{n = 1}^N {\langle {z_{nm}}\rangle },
\end{equation}
\begin{equation}\label{eq:29}
\begin{array}{l}
t_m^* = {t_m} - \sum\limits_{n = 1}^N {\langle {z_{nm}}\rangle [\ln (\sum\limits_{d = 1}^D {{x_{nd}}} ) - \ln (1 + \sum\limits_{d = 1}^D {{x_{nd}}} )]}.
\end{array}
\end{equation}
Takeing the exponential of both sides of \eqref{eq:27}, the equation (30) is obtained as follows
\begin{equation}\label{eq:30}
q({\mathbf{u}}) = \prod\limits_{m = 1}^M {\mathcal G({u_m}|s_m^*,t_m^*)}.
\end{equation}
\emph{4)} $q^{*}(\bf v)$: Considering the derivation of the update equation for the factor $q(\bf v)$, the logarithm of the optimized factor is given by
\begin{equation}\label{eq:31}
\ln q^{*}({v_{m}}) = (p_{m}^* - 1)\ln {v _{m}} - q_{md}^*{v_{m}} + \text{Cst},
\end{equation}
where
\begin{equation}\label{eq:32}
p_m^* = {p_m} + [\Psi ({{\bar u}_m} + {{\bar v}_m}) - \Psi ({{\bar v}_m})]\bar v_{m}\sum\limits_{n = 1}^N {\langle {z_{nm}}\rangle },
\end{equation}
\begin{equation}\label{eq:33}
q_m^* = {q_m} + \sum\limits_{n = 1}^N {\langle {z_{nm}}\rangle \ln (1 + \sum\limits_{d = 1}^D {{x_{nd}}} )}.
\end{equation}
It is obvious that \eqref{eq:31} has a similar form as to the logarithm of the Gamma prior density. Similarly, the equation (34) is obtained as follows
\begin{equation}\label{eq:34}
q^{*}({\bf{v}}) = \prod\limits_{m = 1}^M {\mathcal G({v_m}|p_m^*,q_m^*)}.
\end{equation}
\emph{5)} $q^{*}(\bm\pi)$: Keeping only terms that have a functional dependence on  $\pi_{m}$, the equation (35) is obtained as follows

\begin{equation}\label{eq:35}
\ln {q^*}({\pi _m}) = (c_m^* - 1)\ln {\pi _m} + {\rm{Cst}},
\end{equation}
where
\begin{equation}\label{eq:36}
c_m^* = \sum\limits_{n = 1}^N {\langle {z_{nm}}\rangle }  + {c_m}.
\end{equation}
Takeing the exponential of both sides of \eqref{eq:35}, the equation (37) is obtained as follows
\begin{equation}\label{eq:37}
p(\bm\pi)=Dir(\bm\pi |\bm c^{*})=\frac{{\Gamma (\sum\nolimits_{m = 1}^M {c_m^*} )}}{{\prod\nolimits_{m = 1}^M {\Gamma (c_m^*)} }}\prod\limits_{m = 1}^M {\pi _m^{c_m^* - 1}}.
\end{equation}

All the expected values in the above equations are evaluated by
\begin{equation}\label{eq:38}
 {{\bar \alpha }_{md}} = \frac{{g_{md}^*}}{{h_{md}^*}},\langle \ln {\alpha _{md}}\rangle = \Psi (g_{md}^*) - \ln (h_{md}^*),
\end{equation}
\begin{equation}\label{eq:39}
 {{\bar u }_{m}} = \frac{{s_m^*}}{{t_m^*}},\langle \ln {u_m}\rangle = \Psi (s_m^*) - \ln (t_m^*),
\end{equation}
\begin{equation}\label{eq:40}
 {{\bar v }_{m}} = \frac{{p_m^*}}{{q_m^*}},\langle \ln {v_m}\rangle = \Psi (p_m^*) - \ln (q_m^*),
\end{equation}
\begin{equation}\label{eq:41}
\begin{split}
\langle z_{nm} \rangle = r_{nm}, \langle {\pi _m}\rangle  = \frac{{c_m^*}}{{\sum\nolimits_{m = 1}^M {c_m^*} }}, \langle \ln {\pi _m}\rangle  = \Psi (c_m^*) - \Psi (\sum\limits_{m = 1}^M {c_m^*} ).
\end{split}
\end{equation}
\subsection{Full Variational Learning Algorithm}
With the above obtained variational factors in hand, it is straightforward to evaluate the lower bound \eqref{eq:11} for this model. In practice, it is useful to be able to monitor the bound during the re-estimation in order to test for convergence. The lower bound \eqref{eq:11} is given by
\begin{equation}\label{eq:42}
\begin{split}
\tilde {\cal L}(q) =& \langle \ln \tilde p({\bf{X}},\bm\Theta)\rangle  - \langle \ln q^*(\mathbf Z)\rangle  - \langle \ln q^*(\bm\Lambda )\rangle  - \langle \ln q^*(\mathbf u)\rangle  - \langle \ln q^*(\mathbf v)\rangle -  \langle \ln q^*(\bm\pi)\rangle,\\
\end{split}
\end{equation}
where $\langle \ln \tilde p({\bf{X}},\bm\Theta)\rangle$ is computed using  \eqref{eq:18}. The other terms in the bound are easily evaluated to give the following results:
\begin{equation}\label{eq:43}
\langle \ln {q^*}({\bf{Z}})\rangle  = \sum\limits_{n = 1}^N {\sum\limits_{m = 1}^M {{r_{nm}}\ln {r_{nm}}} },
\end{equation}
\begin{equation}\label{eq:44}
\begin{split}
\langle \ln {q^*}(\bm\Lambda )\rangle  = \sum\limits_{m = 1}^M {\sum\limits_{d = 1}^D {[g_{md}^*\ln h_{md}^* - \ln \Gamma (g_{md}^*)}  + (g_{md}^* - 1)\langle \ln {\alpha _{md}}\rangle  - h_{md}^*\langle {\alpha _{md}}\rangle ]},
\end{split}
\end{equation}
\begin{equation}\label{eq:45}
\begin{split}
\langle \ln {q^*}(\mathbf u )\rangle  = &\sum\limits_{m = 1}^M {[s_m^*\ln t_m^* - \ln \Gamma (s_m^*) + (s_m^* - 1)\langle \ln {u_m}\rangle  - t_m^*\langle {u_m}\rangle ]},
\end{split}
\end{equation}
\begin{equation}\label{eq:46}
\begin{split}
\langle \ln {q^*}(\mathbf v )\rangle  = &\sum\limits_{m = 1}^M {[p_m^*\ln q_m^* - \ln \Gamma (p_m^*) + (p_m^* - 1)\langle \ln {v_m}\rangle  - q_m^*\langle {v_m}\rangle ]},
\end{split}
\end{equation}
\begin{equation}\label{eq:47}
\langle \ln q^{*}(\bm\pi )\rangle  = \ln \frac{{\Gamma (\sum\nolimits_{m = 1}^M {c_m^*} )}}{{\prod\nolimits_{m = 1}^M {\Gamma (c_m^*)} }} + \sum\limits_{m = 1}^M {(c_m^* - 1)\langle \ln {\pi _m}\rangle }.
\end{equation}

The analytically tractable solution for Bayesian estimation of the IBLMM can be obtained in a similar way to the conventional EM algorithm. This inference algorithm is summarized in the Algorithm 1.
\begin{algorithm}
\caption{Algorithm for EVI-based Bayesian IBLMM}
\label{alg:global near-optimal}
\begin{algorithmic}[1]
\STATE  Set the initial values of  $M$, ${{g}_{md}}$, ${{h}_{md}}$, ${{s}_{m}}$, ${{t}_{m}}$,  ${{p}_{m}}$, ${{q}_{m}}$, ${{c}_{m}}$.\\
\STATE Initialize $r_{nm}$ by $K$-Means algorithm.\\
\REPEAT
\STATE The variational E-step: Update $q^{*}(\mathbf Z)$ according to \eqref{eq:21}.
\STATE The variational M-step: Update $q^{*}(\bm\Lambda)$, $q^{*}(\mathbf u)$, $q^{*}(\mathbf v)$ and $q^{*}(\bm\pi)$ according to \eqref{eq:26}, \eqref{eq:30},  \eqref{eq:34}, and \eqref{eq:37}, respectively.
\UNTIL{Stop criterion is reached.}
\STATE  Determine the best number of components $\emph{M}$ via annihilating the components with mixing weights $\pi_{m}\le{10}^{-5}$.
\end{algorithmic}
\end{algorithm}

\section{ Experiments and Results}
In this section, the proposed variational method refered to as EVI-IBLMM is validated through both synthesized datasets and real datasets. The goal of the synthesized dataset validation is to investigate the accuracy of the EVI-IBLMM algorithm in terms of parameter estimation and model selection. The goal of the real dataset validation is to compare the EVI-IBLMM to three other methods: the IDMM applying the EVI technique (EVI-IDMM) \cite{Yuping2018Variational}, the GIDMM applying the EVI technique (EVI-GIDMM) \cite{Ma2019Insights} and the GaMM applying the EVI technique (EVI-GaMM) \cite{Lai2021Extended}. To provide broad noninformative prior distributions,  we set the hyperparameters of the prior distribution as  ${{g}_{md}}={{s}_{m}}={{p}_{m}}=1$, ${{h}_{md}}={{t}_{m}}={{q}_{m}}=0.1$, $c_{m} = 0.001$, and initialize the number of components with
large value (15 in this paper). The initial values of $r_{nm}$ are obtained using the $K$-means algorithm.
Note that this specific selection was based on our experiments and was found to be convenient and effective in our case. When the EVI-IBLMM algorithm stops, the posterior means are taken as the parameter estimates in the IBLMM.

\subsection{Synthesized Data Validation}
The performance of the proposed EVI-IBLMM in terms of estimation and determination through quantitative analysis on four 2-D synthesized datasets is first evaluated, which are generated from four known IBLMMs with different parameters. It is worth noting that the selection of $D = 2$ is purely for ease of representation. Table 1 shows the actual parameters for the four IBLMMs. The initial number of components for each dataset are set to double amounts of the actual number of components with equal mixture weights.
\linespread{0.8}
\begin{table*}[!htbp]\centering
\renewcommand{\arraystretch}{1.1}
\caption{True values of the parameters in the IBLMM applied to generate the four synthesized datasets.}
\label{Table1}
\centering
\tabcolsep 0.07in
\begin{tabular}{cc|ccccc}
\hline
Dataset& $m$ & $\alpha_{m1}$ & $\alpha_{m2}$ & $u_{m}$ &  $v_{m}$  &$\pi_m$  \\
\hline
\multirow{2}*{A} & 1 & 12.00 & 24.00 & 8.50 & 12.50 & 0.400  \\
   & 2 & 21.00 & 15.00 & 18.00 & 5.00 & 0.600 \\
\hline
\multirow{3}*{B} & 1 & 12.00 & 24.00 & 8.50 & 12.50  & 0.200  \\
   & 2 & 21.00 & 15.00 & 18.00 & 5.00  & 0.300  \\
   & 3 & 18.50 & 8.00 & 4.00  & 16.50 & 0.500 \\
   \hline
\multirow{4}*{C}  & 1 & 12.00 & 21.00 & 8.50& 12.50 & 0.100  \\
   & 2 & 21.00 & 35.00 & 18.00 & 5.00  & 0.200 \\
   & 3 & 32.00 & 28.00 & 4.00  & 16.50  & 0.300 \\
    & 4 & 2.00 & 18.00 & 24.00 &8.00& 0.400\\
\hline
\multirow{5}*{D} & 1 & 21.00 & 6.00 & 18.00 & 24.00 & 0.100 \\
   & 2 & 2.00 & 28.00 & 8.00 & 15.00  & 0.200 \\
   & 3 & 18.00 & 68.00 & 24.00  & 16.00  & 0.250  \\
   & 4 & 76.00 & 8.00 & 4.00 &18.00 & 0.300\\
   & 5 & 2.00 & 4.00 & 4.00  & 12.00& 0.150 \\
\hline
\end{tabular}
\end{table*}
The average estimated parameters of the four generated datasets over 20 runs of simulations are reported in Table 2. According to these results, the proposed EVI-IBLMM algorithm is capable of accurately estimating both the parameters and the mixing weights of the IBLMM.
\linespread{0.8}
\begin{table*}[!htbp]\centering
\renewcommand{\arraystretch}{1.1}
\caption{The mean of the estimated parameters for the synthesized datasets over 20 runs of the EVI-IBLMM algorithm.}
\label{Table1}
\centering
\tabcolsep 0.07in
\begin{tabular}{ccc|ccccc}
\hline
Dataset& $N_{m}$& $m$ & $\hat\alpha_{m1}$ & $\hat\alpha_{m2}$ & $\hat u_{m}$ &  $\hat v_{m}$  &$\hat\pi_m$   \\
\hline
\multirow{2}*{A} &200 & 1 & 11.99 & 23.95 & 8.56 & 12.51 & 0.400  \\
   &300 & 2 & 21.27 & 15.20 & 18.10 & 5.00 & 0.600\\
\hline
\multirow{3}*{B} &120 & 1 & 11.31 & 22.59 & 8.50 & 12.54 & 0.200 \\
   & 180& 2 & 20.81 & 14.93 & 18.50 & 5.13 & 0.300 \\
   & 300& 3 & 18.30 & 8.01 & 4.18  & 17.09 & 0.500  \\
   \hline
\multirow{4}*{C}  & 80& 1 & 12.46 & 21.64 & 9.20& 14.12 & 0.098\\
   & 160& 2 & 19.84 & 33.52 & 18.30 & 5.08 & 0.202 \\
   &240 & 3 & 30.68 & 26.81 & 4.07  & 16.76 & 0.300\\
    &320 & 4 & 2.00 & 18.12 & 24.32 &8.21 &0.400 \\
\hline
\multirow{5}*{D} & 100& 1 & 22.26 & 6.42 & 17.70 & 23.46 & 0.103 \\
   &200 & 2 & 1.98 & 27.09 & 7.80 & 15.03 & 0.201 \\
   &250 & 3 & 16.61 & 64.69 & 23.79  & 15.82 & 0.253 \\
   &300 & 4 & 73.02 & 7.48 & 4.04 &18.10 &0.302 \\
   & 150& 5 & 2.32 & 4.14 & 3.98  & 12.11 &0.141  \\
\hline
\end{tabular}
\end{table*}
Next, the model selection capability of the EVI-IBLMM algorithm is investigated. When the initial number of components is larger than the true one, the EVI-IBLMM algorithm is capable of forcing some of the mixing weights to approach zero. These components make little contribution to the model, thus they can be eliminated. The EVI-IBLMM algorithm is initiated with a mixture of many components (15 in this paper) and equal mixture weights. Figure 1 shows the estimated mixture weights of each component for the different generated datasets after convergence. According to these results, it can be clearly observed that the EVI-IBLMM algorithm is able to effectively determine the model complexity.
\begin{figure}[!htp]
 \centering
 \begin{tabular}{cc}
   \includegraphics[width=7.0cm]{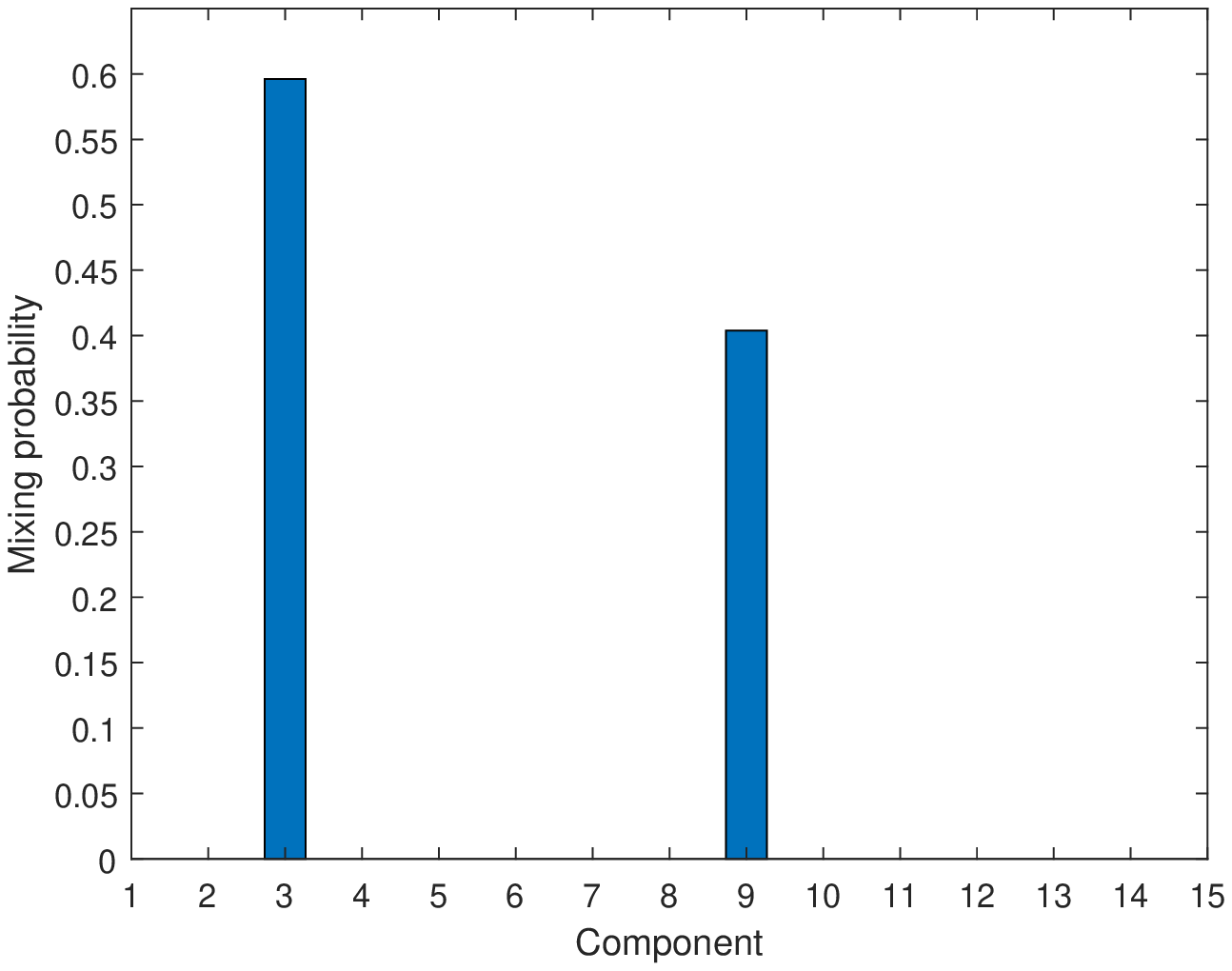} & \includegraphics[width=7.0cm]{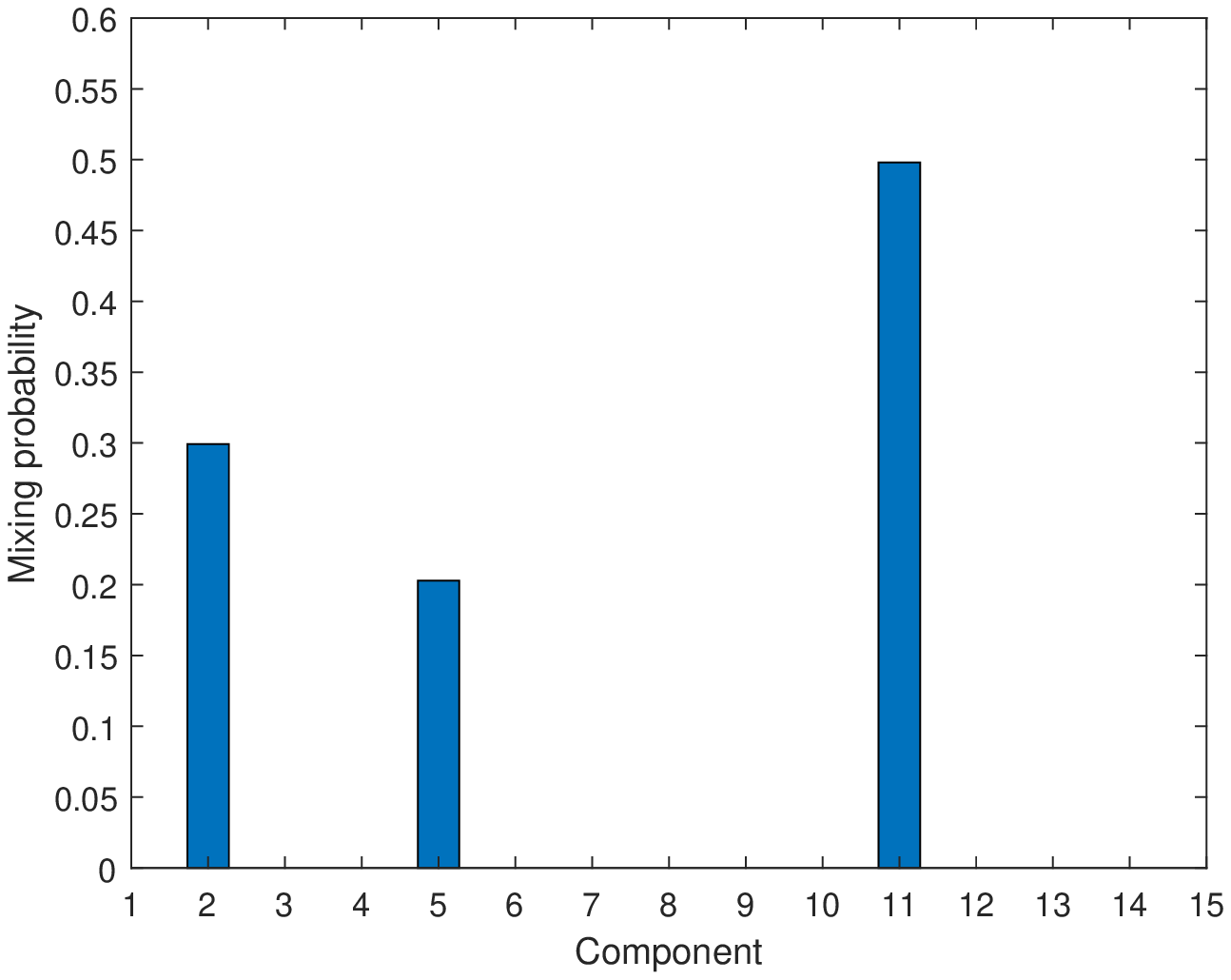} \\
   (a)  & (b)   \\
    \includegraphics[width=7.0cm]{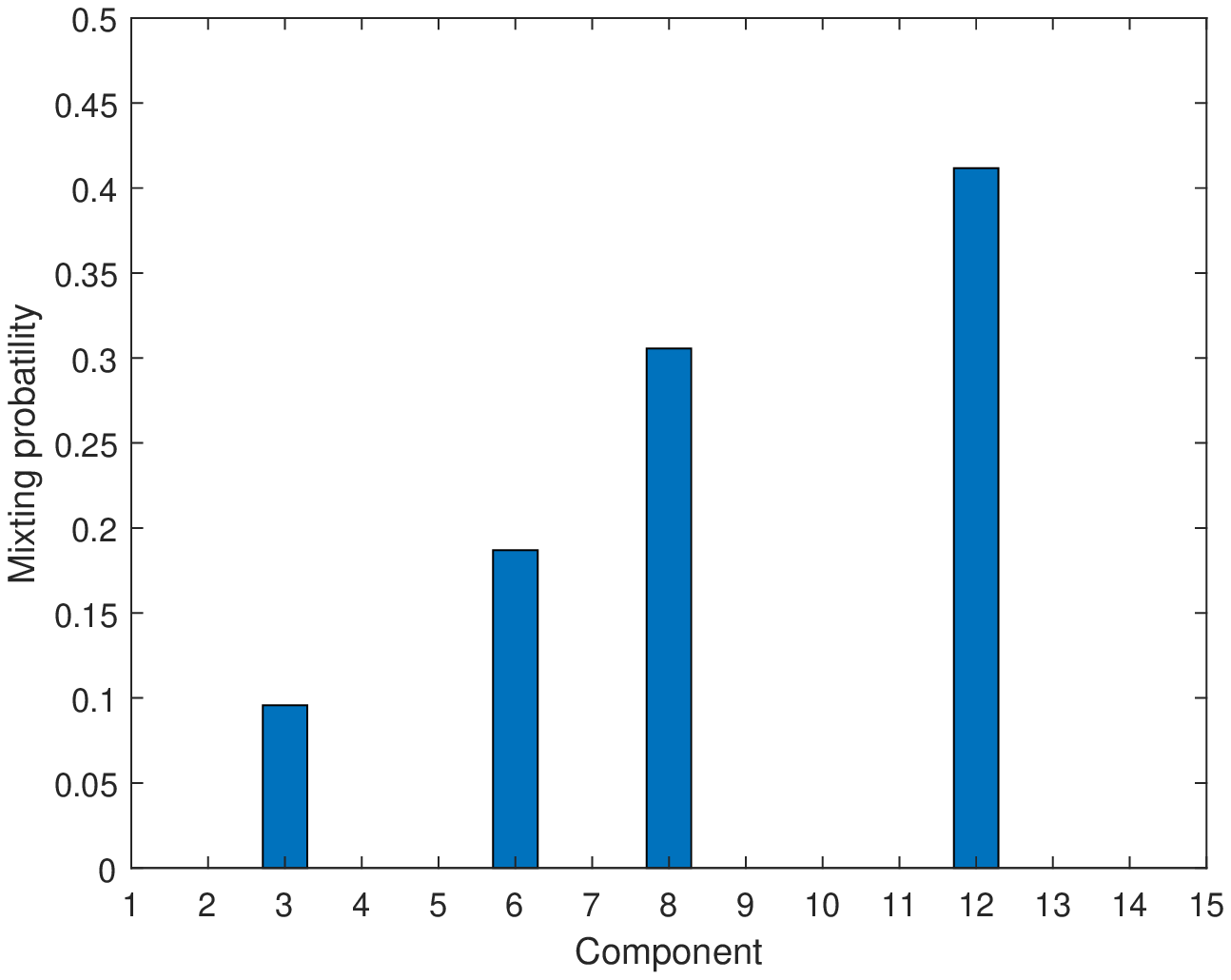} & \includegraphics[width=7.0cm]{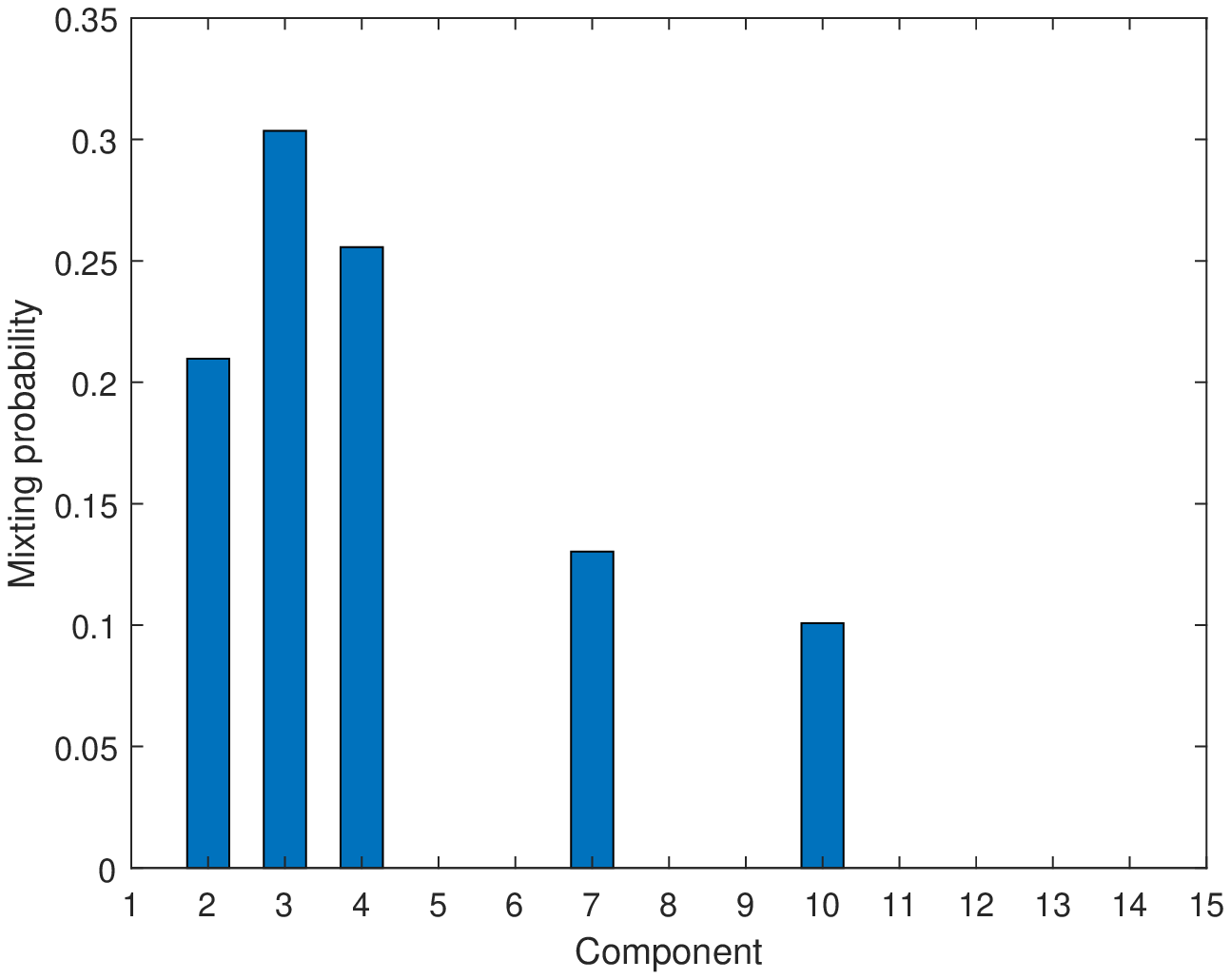}\\
    (c)  & (d) \\
 \end{tabular}
\caption{\label{fig2} Estimated mixing probabilities of components  for the synthesized datasets. (a) Dataset A. (b) Dataset B. (c) Dataset C. (d) Dataset D.}
\end{figure}
Then, the effect of initial number of components upon the resulting model complexity is investigated. Based on datset A, Figure 2 shows the effect of initial number of components on the resulting model complexity over 100 runs of simulations.  According to the results shown in this picture, the EVI-IBLMM algorithm is capable of identifying the accurate number of components regardless of whether the sample size is small or large. Moreover, as the sample size gets larger, the effect of the initial number of components gets more insignificant.
Finally, the convergence of the  EVI-IBLMM algorithm is investigated.
\begin{figure}[!htp]
 \centering
 \begin{tabular}{cc}
   \includegraphics[width=7.0cm]{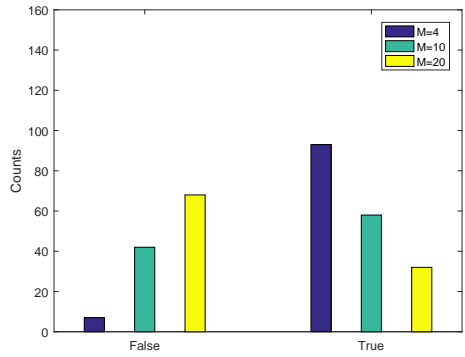} & \includegraphics[width=7.0cm]{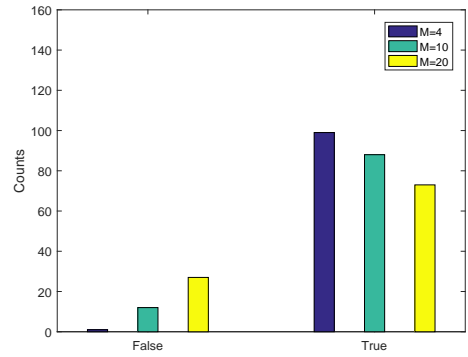} \\
   (a) N=50 & (b) N=100  \\
    \includegraphics[width=7.0cm]{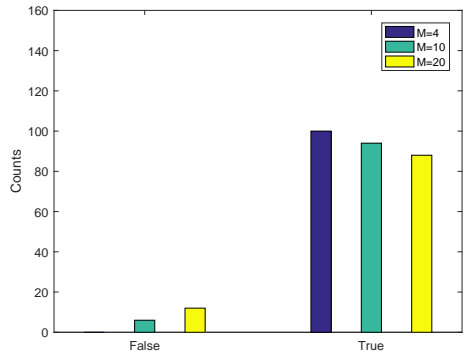} & \includegraphics[width=7.0cm]{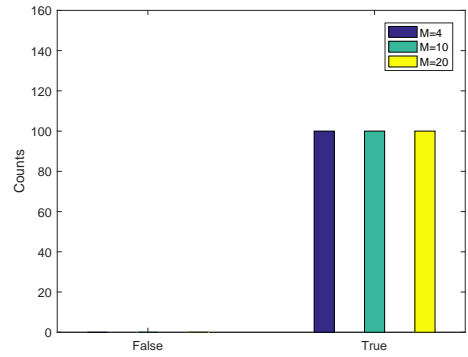}\\
    (c) N=200 & (d) N=500 \\
 \end{tabular}
\caption{\label{fig2} The counts of the estimated number of components over 100 runs of simulations based on dataset A. $M$ denotes the initial number of components and $N$ denotes the sample size.}
\end{figure}
Figure 3 shows the value of the variational objective function in each iteration. According to this figure, it is clear that the variational objective function is always increasing during iterations, thus the convergence is demonstrated.
\begin{figure}[!htp]
 \centering
 \begin{tabular}{cc}
   \includegraphics[width=7.0cm]{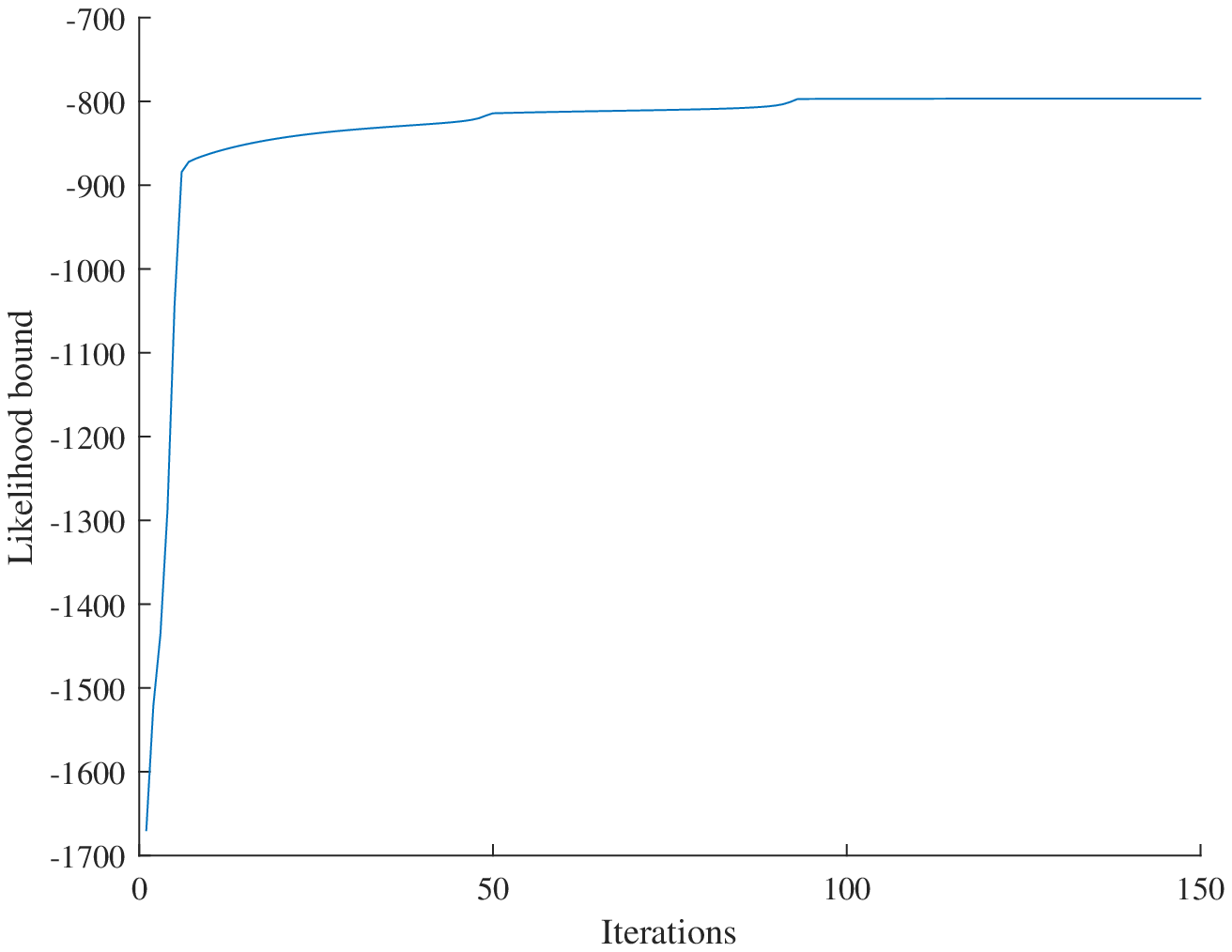} & \includegraphics[width=7.0cm]{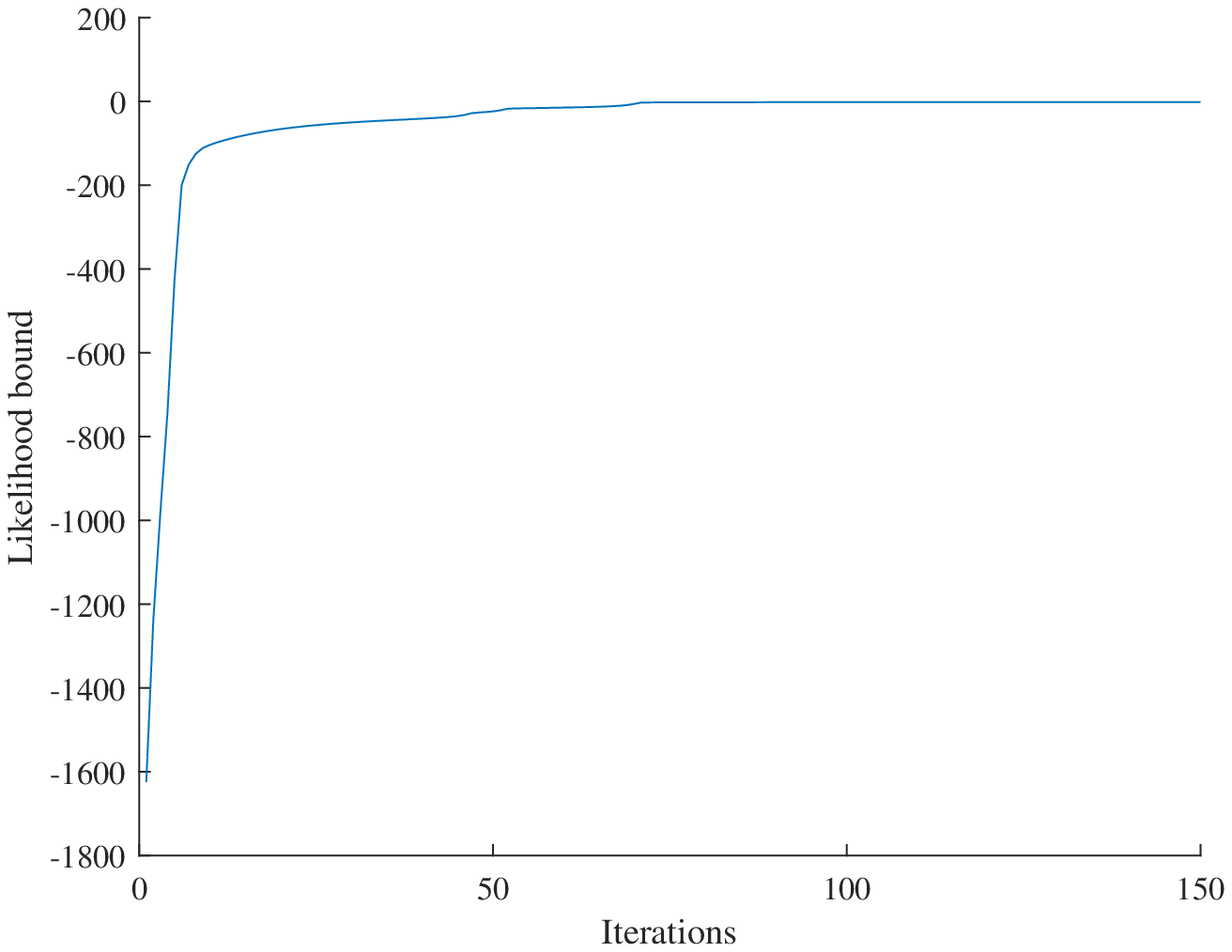} \\
   (a) & (b)  \\
    \includegraphics[width=7.0cm]{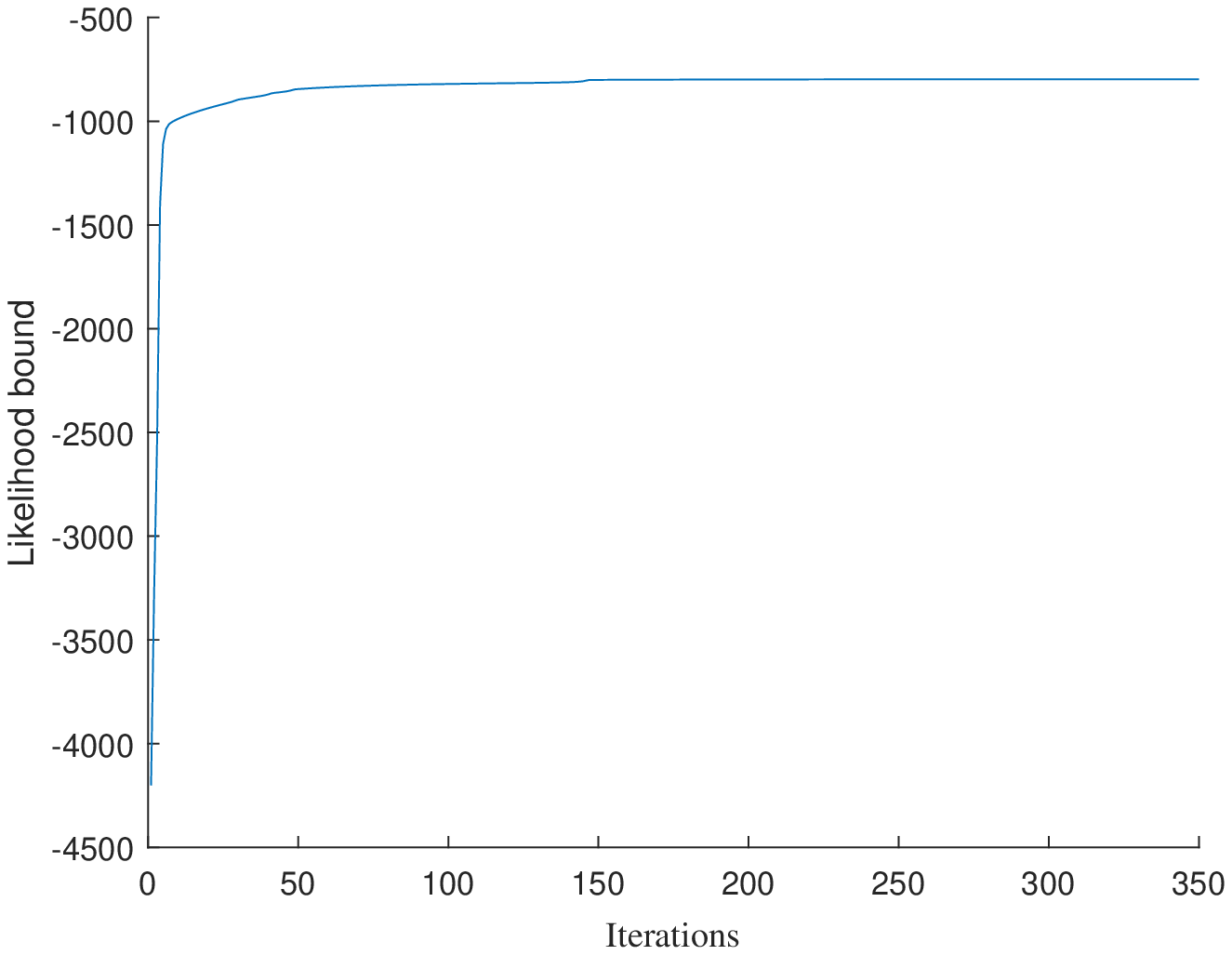} & \includegraphics[width=7.0cm]{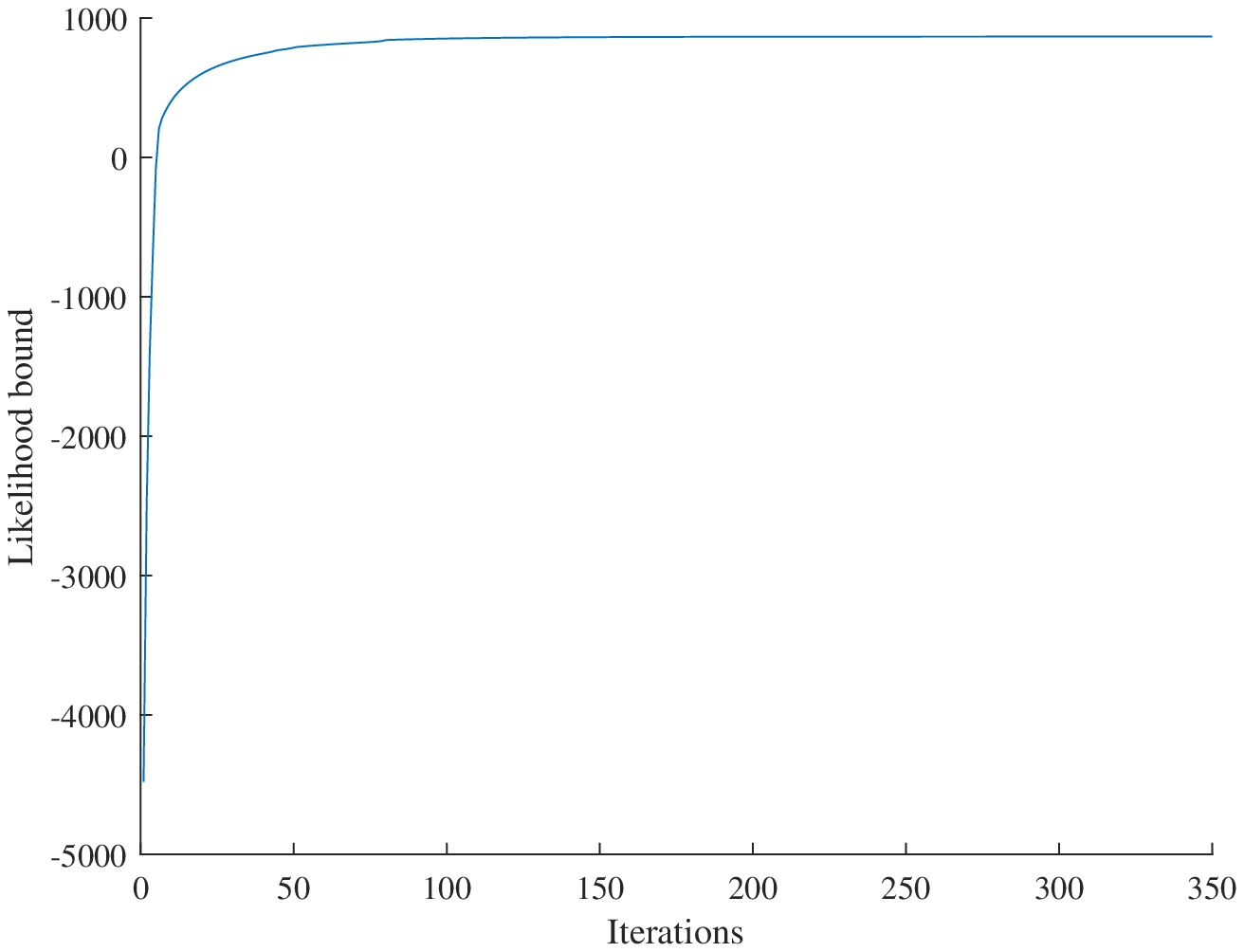}\\
    (c)  & (d) \\
 \end{tabular}
\caption{\label{fig2} Convergence of the proposed EVI-IBLMM algorithm for the different synthesized datasets. (a) Dataset A. (b) Dataset B. (c) Dataset C. (d) Dataset D. }
\end{figure}

\subsection{Text categorization}
Text categorization refers to the task of automatically assigning unlabeled text documents into predefined categories. During the past few decades, this task has attracted considerable attention from researchers due to many reasons, such as the hug amount of digital documents that are easily available and the increasing demand to organize, store, and retrieve these documents accurately and efficiently. Efficient text categorization are beneficial for many applications, such as document processing and visualization \cite{Strobelt2009Document}, digital information search \cite{Chu2000A}, and information retrieval \cite{Peng2019A}. This problem is challenging and different statistical methods were proposed and applied in the past. Although different, most of the proposed techniques addressed this problem as following: First, a set of labeled text documents which belong to a certain number of classes are given to train the model; Second, a new unobserved text is assigned to the category with the highest similarity regarding its content by the model.

The text categorization experiment with the proposed EVI-IBLMM in our paper is conducted by using two extensively applied text collections: WebKB \cite{Nigam1998Learning} and 20Newsgroup \footnote{ http://kdd.ics.uci.edu/databases/20news
groups/20newsgroups.html}. The WebKB dataset  is composed of four categories: course, faculty, project and student, with a total of 4,199 documents. The 20Newsgroups dataset contains 13,998 newsgroup documents evenly distributed on 20 categories. Each of these categories is 30 times randomly divided into two separate halves, one half for training and the other half for testing. Following \cite{Ping2012Efficient}, the Porter's stemming \cite{Porter1980An} is applied to reduce the words to their basic forms. In the pre-processing step, the words that occur less than 3 times or is shorter than 2 in length are eliminated, which results in the representation of each document by a positive vector. The vectors in the different training sets are then modeled by the IBLMM trained by the algorithm in the previous section. Finally, each document vector is categorized to a given category according to the well-known Bayes classification rule.

Three referred  methods, namely the EVI-based Bayesian GIDMM \cite{Ma2019Insights} (EVI-GIDMM), EVI-based Bayesian IDMM (EVI-IDMM) \cite{Yuping2018Variational}  and  EVI-based Bayesian Gamma mixture model (EVI-GaMM)  \cite{Lai2021Extended} are also used to the aforementioned task. Table 3 shows the mean results of the tested methods in terms of categorization accuracy and training time over 20 runs. Figure 5 illustrates the categorization accuracies obtained by different methods. Based on these results, it can be found that the proposed EVI-IBLMM has the best categorization accuracy ($\%$) among all the referred mixture-based approach for the task of text categorization.
\begin{table}[!htp]
\centering
\tabcolsep 0.10in
\caption{Comparisons of text categorization accuracies (in $\%$) and runtime (in s) obtained by different  approaches.}
\begin{tabular}{c||c||cccc}
\hline
Dataset & Method  & EVI-IBLMM & EVI-GIDMM& EVI-IDMM& EVI-GaMM \\
\hline
\multirow{2}*{WebKB}  & Accuracy  & \textbf{90.36} &  89.27 &  89.91 &89.03 \\

~ & Runtime  &     0.66   &   0.61   &   0.59    &   0.39\\
\hline
\multirow{2}*{20Newsgroup} &  Accuracy  & \textbf{81.11} &  79.82 & 80.20 & 78.86 \\

~ & Runtime  &  4.85    &   5.35    &   3.84    &    0.71    \\
\hline
\end{tabular}
\end{table}
\begin{figure}[!htp]
 \centering
 \begin{tabular}{cc}
   \includegraphics[width=7.0cm]{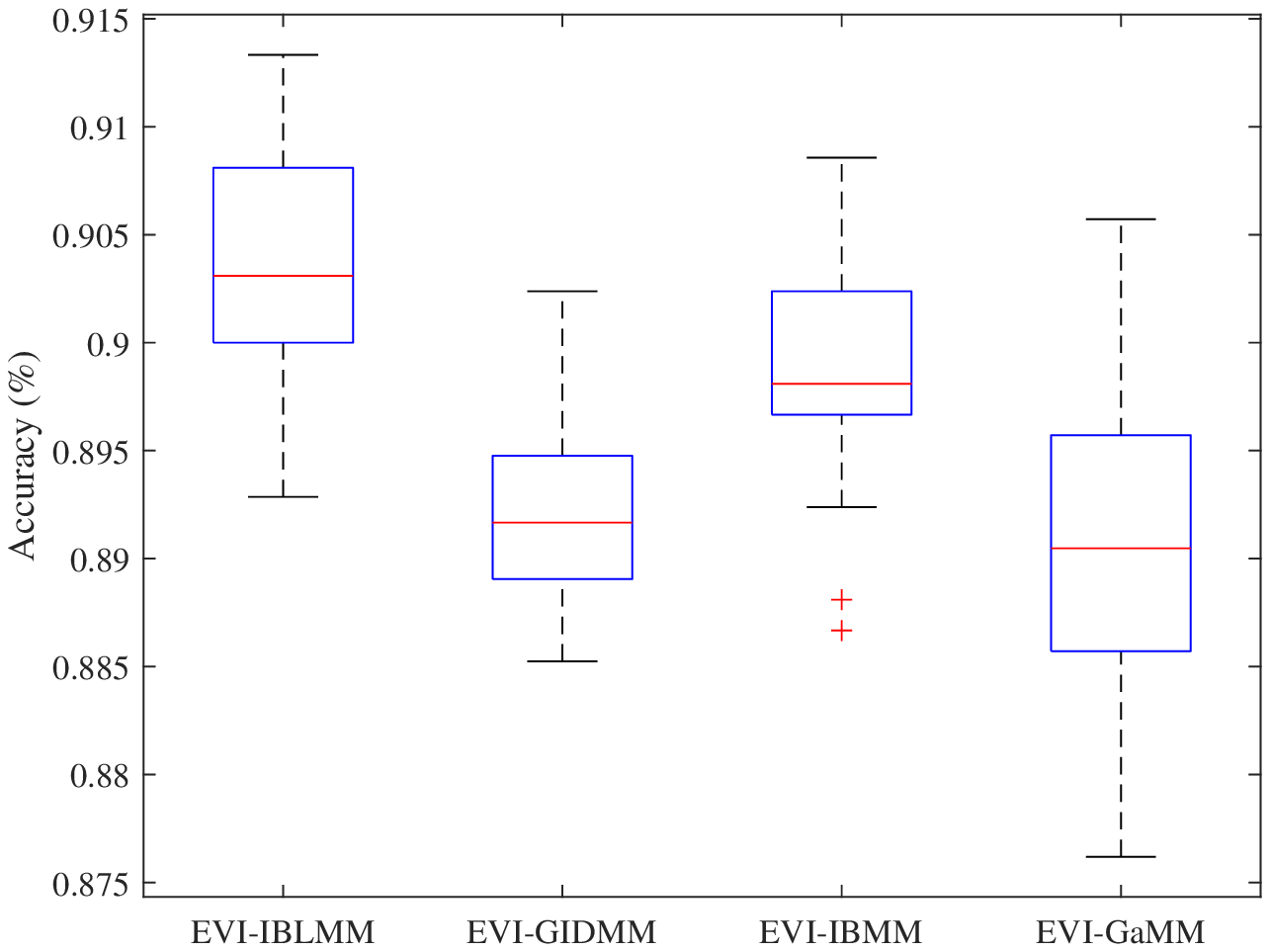} & \includegraphics[width=7.0cm]{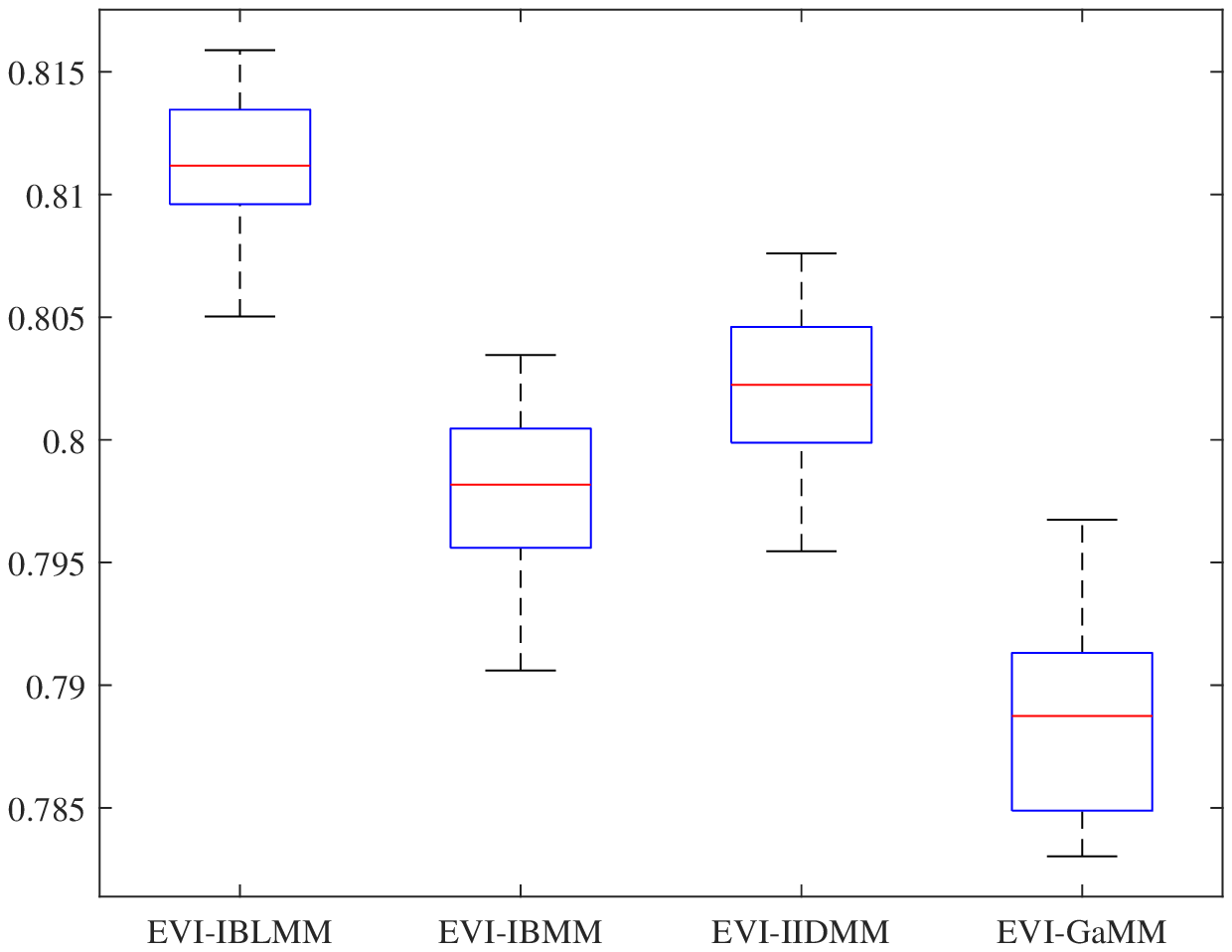} \\
   (a) WebKB  & (b)  20Newsgroup \\
 \end{tabular}
\caption{\label{fig2} Boxplots for comparisons of the categorization accuracies' distributions for the WebKB and the 20Newsgroup datasets.}.
\end{figure}
Moreover, to investigate more insights for the EVI-IBLMM algorithm, the EVI-IBLMM is further compared with deep neural networks (DNNs) on the text categorization task. The fully connected (FC) neural networks with different numbers (\emph{i.e.}, $l$) of hidden layers are used. The extracted feature vectors for the WebKB and 20Newsgroup datasets are used as inputs, respectively. These feature vectors are named as shallow feature vectors. The $l$ is set as 1, 2, and 4, respectively and the number of nodes in each hidden layer is the same as the dimension of the shallow features. Table 4 shows the comparison of categorization accuracies and training time of different FC neural networks and the proposed EVI-IBLMM algorithm on both WebKB and 20Newsgroup datasets.  According to these results, it can be found that the proposed method significantly decreases training time than the FC neural networks. Although the proposed approach cannot outperform the DNNs, it can effectively model the features extracted and obtain proper classification accuracies on the two datasets, which can explicitly show the effectiveness of the proposed method.

\begin{table}[!htp]
\centering
\tabcolsep 0.10in
\caption{Comparisons of text categorization accuracies (in $\%$) and runtime (in s) obtained by different approaches. Note that $l$ means number of hidden layers of the FC neural networks.}
\begin{tabular}{c||c||cccc}
\hline
Dataset & Method  &  {FC ($l$=1)}  & {FC ($l$=2)}  &  {FC ($l$=4)} & EVI-IBLMM \\
\hline
 \multirow{2}*{WebKB} & Accuracy  & 89.64 & 87.06 & 83.40 &  \textbf{90.36}  \\

~ & Runtime  & 4.77  &   6.68   &    6.91   &  \textbf{ 0.66 }  \\
\hline
\multirow{2}*{20Newsgroup} &  Accuracy  & 81.39   & 81.29  &   81.06  &  \textbf{81.11} \\

& Runtime  &   16.39   &    19.03   &   43.72    & \textbf{ 4.85 }  \\
\hline
\end{tabular}
\end{table}

\section{Conclusions}
In this paper, an efficient attractive EVI algorithm for the inverted Beta-Liouville mixture model is proposed. This algorithm is able to automatically and simultaneously determine all the model's parameters and the optimal number of components, which can prevent the problem of over-fitting. The good performance of the proposed method are experimentally demonstrated through both synthetic datasets and real datasets which are generated from a real-world application namely text categorization. A future work can be devoted to investigate how to combine a feature selection criterion with the model selection in a unified Bayesian framework or to extend the IBLMM to the infinite case applying some nonparametric Bayesian methods.
\section*{Acknowledgment}
This work was supported by the General Project of Science and Technology Plan of Beijing Municipal Commission of Education (No. KM201910009014)
and the National Natural Science Foundation of China (Grant No. 62172193).

\section*{References}
\bibliographystyle{elsarticle-num}
\bibliography{myreference}

\end{document}